%% file: iclr2023_conference.tex
\documentclass{article} % For LaTeX2e
\usepackage{iclr2023_conference,times}

\input{math_commands.tex}

\usepackage{hyperref}
\usepackage{url}
\usepackage{graphicx}

\usepackage{algorithm}
\usepackage{algorithmic}
\usepackage{enumitem}
\usepackage{xcolor}

\usepackage{float}
\usepackage{booktabs}
\usepackage{appendix}
\usepackage{listings}
\usepackage{authblk}

\title{Let's reward step by step: \\ Step-Level reward model as the Navigators for  Reasoning }

\author[1]{\textbf{Qianli Ma}}
\author[1]{\textbf{Haotian Zhou}}
\author[2]{\textbf{Tingkai Liu}}
\author[2]{\textbf{Jianbo Yuan}}  
\author[3,\dag]{\\ \textbf{Pengfei Liu}}
\author[1,\dag]{\textbf{Yang You}}
\author[2,\dag]{\textbf{Hongxia Yang}}

\affil[1]{School of Computing, National University of Singapore}
\affil[2]{ByteDance, Inc.}
\affil[3]{Generative Artificial Intelligence Research Lab, Shanghai Jiao Tong University}
\affil[$\dag$]{Corresponding authors}

\definecolor{positive}{HTML}{E5F6DA}
\definecolor{neutral}{HTML}{FBEADA}
\definecolor{negative}{HTML}{FADCDB}

\iclrfinalcopy % Uncomment for camera-ready version, but NOT for submission.
\begin{document}

\maketitle
\input{abstract}

\input{introduction}

\input{method}

\input{experiment}

\input{related}

\input{conclusion}

\newpage

\section*{Acknowledgement}
Yang You's research group is being sponsored by NUS startup grant (Presidential Young Professorship), Singapore MOE Tier-1 grant, ByteDance grant, ARCTIC grant, SMI grant Alibaba grant, and Google grant for TPU usage.

\bibliography{iclr2023_conference}
\bibliographystyle{iclr2023_conference}

\newpage
\appendix
\input{appendix}

\end{document}

%% file: math_commands.tex
\usepackage{amsmath,amsfonts,bm}

\def\eqref#1{equation~\ref{#1}}
\def\1{\bm{1}}

\DeclareMathAlphabet{\mathsfit}{\encodingdefault}{\sfdefault}{m}{sl}
\SetMathAlphabet{\mathsfit}{bold}{\encodingdefault}{\sfdefault}{bx}{n}

%% file: abstract.tex
\begin{abstract}
Recent years have seen considerable advancements in multi-step reasoning with Large Language Models (LLMs). The previous studies have elucidated the merits of integrating feedback or search mechanisms during model inference to improve the reasoning accuracy. The Process-Supervised Reward Model (PRM), typically furnishes LLMs with step-by-step feedback during the training phase, akin to Proximal Policy Optimization (PPO) or reject sampling. Our objective is to examine the efficacy of PRM in the inference phase to help discern the optimal solution paths for multi-step tasks such as mathematical reasoning and code generation. To this end, we propose a heuristic greedy search algorithm that employs the step-level feedback from PRM to optimize the reasoning pathways explored by LLMs. This tailored PRM demonstrated enhanced results compared to the Chain of Thought (CoT) on mathematical benchmarks like GSM8K and MATH. Additionally, to explore the versatility of our approach, we develop a novel method to automatically generate step-level reward dataset for coding tasks and observed similar improved performance in the code generation tasks. Thus highlighting the robust nature of our reward-model-based approach to inference for reasoning tasks.
\end{abstract}

%% file: introduction.tex
\section{Introduction}

In the exciting evolution of Large Language Models (LLMs) such as GPT \citep{openai2023gpt4, brown2020language}, LLaMA \citep{touvron2023llama, touvron2023llama2}, OPT \citep{zhang2022opt}, Falcon \citep{refinedweb}, and PaLM \citep{anil2023palm, chowdhery2022palm}, a consistent ability to handle tasks from conversation to text generation has been evident. However, when it comes to reasoning, especially multi-step reasoning, current LLMs, even with sophisticated prompting techniques like the Chain of Thought (CoT)\citep{wei2023chainofthought}, are still prone to a cascade of errors in their generation processes. As the number of reasoning steps increases, these LLMs face challenges in providing and integrating effective feedback, resulting in one error leading to another.

Achieving a refined multi-step reasoning capability for LLMs can unlock their potential across an even broader array of applications, ranging from complex problem-solving to high-level intellectual tasks. Furthermore, as these models become foundational tools in numerous domains, ensuring their reasoning accuracy is paramount to prevent compounding errors and instill trust in their outputs.

To address the challenges of multi-step reasoning, prior research has focused on feedback mechanisms and supervisory techniques. Works have incorporated self-reflection \citep{madaan2023selfrefine, xie2023decomposition} or external observation \citep{yao2023react, gou2023critic} during reasoning, aiming to correct errors and determine correct reasoning steps. Additionally, novel methods like Tree of Thought (ToT)\citep{yao2023tree} and RAP\citep{hao2023reasoning} have integrated traditional search algorithms to convert multi-step reasoning into a path search problem. However, while some of these solutions have led to improvements, they introduce new sets of problems. For instance, the risk of LLMs falling into repetitive thought loops or their self-reflection process exceeding the model's context window constraints. Additionally, direct search processes, such as BFS or DFS, often result in redundancy and substantial reasoning overhead, which, in extreme cases, can degrade to exhaustive search complexity.

Building upon previous studies, it's found that enhancing the reasoning capabilities of large language models involves a few essential aspects. These include step-by-step reasoning, which helps in breaking down complex problems incrementally and guiding the model toward the solution. Another crucial element is the incorporation of feedback or self-reflection. Such feedback can be from the model itself, other models, or even external sources like the results of code execution. Additionally, implementing heuristic search algorithms can be beneficial. Especially when trying to facilitate the model's reasoning path search, it becomes important to compress the search space effectively.

\cite{uesato2022solving} and \cite{lightman2023lets} introduced the process-supervised reward model (PRM). PRM can provide step-level feedback for the multi-step reasoning result generated by the language model, but it has traditionally been used in RLHF \citep{NEURIPS2022_b1efde53, luo2023wizardmath} via proximal policy optimization (PPO) \citep{schulman2017proximal} or reject sampling \citep{yuan2023rrhf, dong2023raft, bai2022constitutional}. Thus, we hypothesized that the fine-grained feedback from PRM, which enhances RLHF \citep{wu2023finegrained}, could similarly improve reasoning path searching. PRM was used as a reward model for the PPO process in \cite{luo2023wizardmath}. Ideally, PRM would not introduce additional training overhead and could be further utilized in the decoding phase after the PPO stage to enhance the model's reasoning capabilities. On the other hand, PRM only requires evaluation for each step, making its computational cost much lower than model self-reflection.

Consequently, we propose a heuristic greedy search algorithm for large language model reasoning tasks that harnesses the process-supervised reward model for step-level feedback(HGS-PRM). Similar to the \cite{lightman2023lets}, we trained a process-supervised reward model based on the open-source model. With our method, each new reasoning step generated by the large language model undergoes evaluation by the reward model. This determines whether to accept the given step or continue generating a new one. If a valid new step is unattainable, a backtrack occurs until a complete reasoning path is identified. Our experimental results on GSM8K \citep{cobbe2021training} and MATH \citep{hendrycksmath2021} both surpassed those achieved with the CoT method. For instance, with the WizardMath-13B \citep{luo2023wizardmath}model, our method achieved accuracies of 65.4\% and 13.7\% on the GSM8K and MATH datasets, respectively, surpassing CoT's 63.2\% and 10.4\%.

Beyond mathematical problems, we aimed to demonstrate the efficacy of HGS-PRM in code generation and even more intricate reasoning tasks. By leveraging technologies like Abstract Syntax Trees (AST) and Mutation Testing, we automatically generated PRM data for code generation challenges on the MBPP \citep{austin2021program} dataset. Following this, we trained a process-supervised reward model specifically tailored for code. Moreover, our approach also yielded improved results on the HumanEval \citep{chen2021codex} benchmark. For instance, the results for Code-LLaMA-Python-13B increased from 41.5\% to 44.5\%.

In summary, our contributions are as follows:

\begin{enumerate}[leftmargin=*]
    \item We explored the feasibility of utilizing PRM during the decoding phase. Training and deploying PRM is a challenging task. By demonstrating the viability of PRM-assisted decoding on open-source models, we've achieved a significant advancement.
    \item We evaluated the strengths and weaknesses of prior methods for enhancing large language model reasoning capabilities. Innovatively, we combined PRM with path search, introducing a heuristic greedy search algorithm that leverages PRM for step-level feedback.
    \item Ours is the pioneering effort to apply PRM in the coding domain. We have released a PRM dataset specifically for code and validated that PRM is effective across a broader spectrum of reasoning tasks.
\end{enumerate}

\begin{figure}[h]
    \centering
    \includegraphics[width=0.7\linewidth]{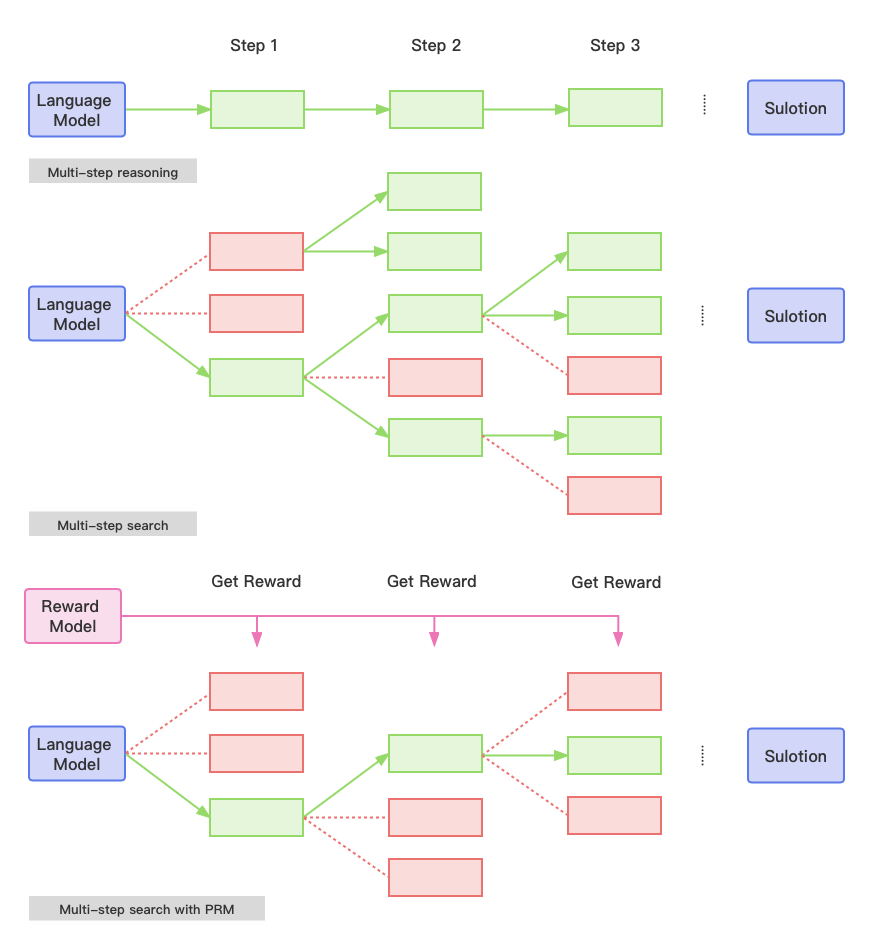}
    \caption{The illustrations of the differences among multi-step reasoning, multi-step search, and multi-step search with PRM}
    \label{fig: RM search}
\end{figure}

%% file: method.tex
\section{Method}
\label{gen_inst}

In this section, we delve into the technical methodologies we have adopted. First and foremost, we discussed how to use step-level data to train our process-supervised reward model(PRM). Moreover, we provide an in-depth examination of the heuristic greedy search algorithm with PRM (HGS-PRM), discussing its operational mechanics and core concepts. To demonstrate the versatility of this method for other reasoning tasks, We will also introduce how we generated our code-specific PRM data by employing Abstract Syntax Tree (AST) techniques to automate the generation of PRM data tailored for code-related tasks.

\subsection{Process-supervised reward model}
Following \cite{alpaca} and \cite{lightman2023lets}, we trained our reward model, with the entire process divided into two phases:

\begin{enumerate}[leftmargin=*]
    \item To equip the foundational LLaMA model with mathematical instruction capabilities and strengthen its mathematical proficiency, we initially used the training portion of the MATH dataset. In line with \cite{alpaca} approach, we mapped 'question' and 'solution' from the MATH\citep{hendrycksmath2021} training dataset to 'instruction' and 'response', respectively. We then performed instruction tuning\citep{wang-etal-2023-self-instruct} on the LLaMA-7B\citep{touvron2023llama} model and got LLaMA-7B-SFT.
    \item Building on the instruction-tuning basis, we trained a reward model using PRM800k\citep{lightman2023lets} based on LLaMA-7B-SFT. This model is designed to score each step of the Solution on a step-by-step dimension. The labels fall into three categories: Positive, Negative, and Neutral.
\end{enumerate}

\subsection{Heuristic Greedy Search with PRM}

Heuristic search is an algorithmic strategy that utilizes heuristic information in problem-solving to guide the search direction, thereby finding solutions more rapidly. A classic example of heuristic greedy search is the A* algorithm, which combines the benefits of best-first search and Dijkstra's algorithm. It efficiently finds the shortest path by using a heuristic to estimate the cost to the goal from a given node.

During the mathematical problem-solving process with large language models, each chain of reasoning can be broken down step by step. This can be translated into a search problem for the optimal reasoning path within a vast space. Assessing the reward of each step, or the value of a specific state, is crucial for enhancing the inference capabilities of large language models through search algorithms. 

Utilizing the language model's self-evaluation to judge the value of a step can lead to significant computational overhead. For autoregressive models like GPT\citep{radford2019language} that rely solely on decoding, the computational time complexity is \(O(N^2)\). The introduction of self-assessment lengthens the sequence, causing the time complexity to grow exponentially. Additionally, traditional search methods like BFS and DFS have an expansive search space and involve many redundant nodes.

In comparison, using only a process-supervised reward model for assessment significantly reduces overhead compared to self-assessment. Moreover, heuristic greedy search algorithms can effectively minimize the search space, yielding faster results.

\begin{figure}[h]
    \centering
    \includegraphics[width=0.7\linewidth]{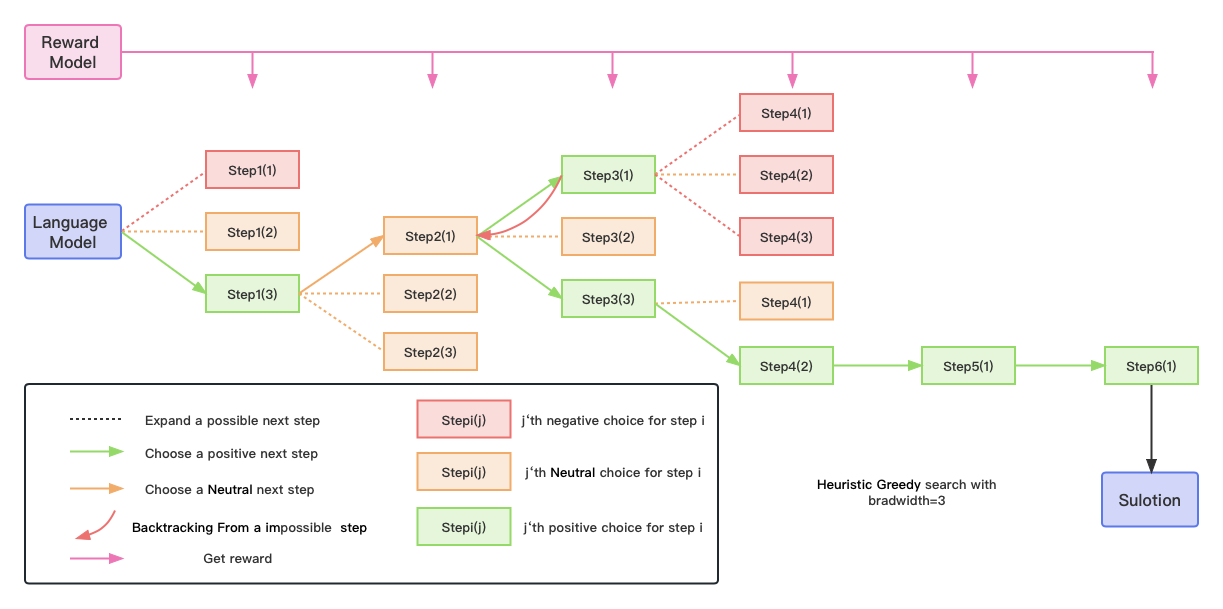}
    \caption{The illustration above depicts the workflow of the heuristic greedy search assisted by the process-supervised reward model, encompassing the expansion of the next step, scoring by the reward model, backtracking, and pinpointing the final result. \colorbox{negative}{Red} denotes a negative step, \colorbox{positive}{Green} indicates a positive step, and \colorbox{neutral}{Orange} signifies a neutral step.}
    \label{fig: greedy search}
\end{figure}

\begin{figure}[h]
    \centering
    \includegraphics[width=0.7\linewidth]{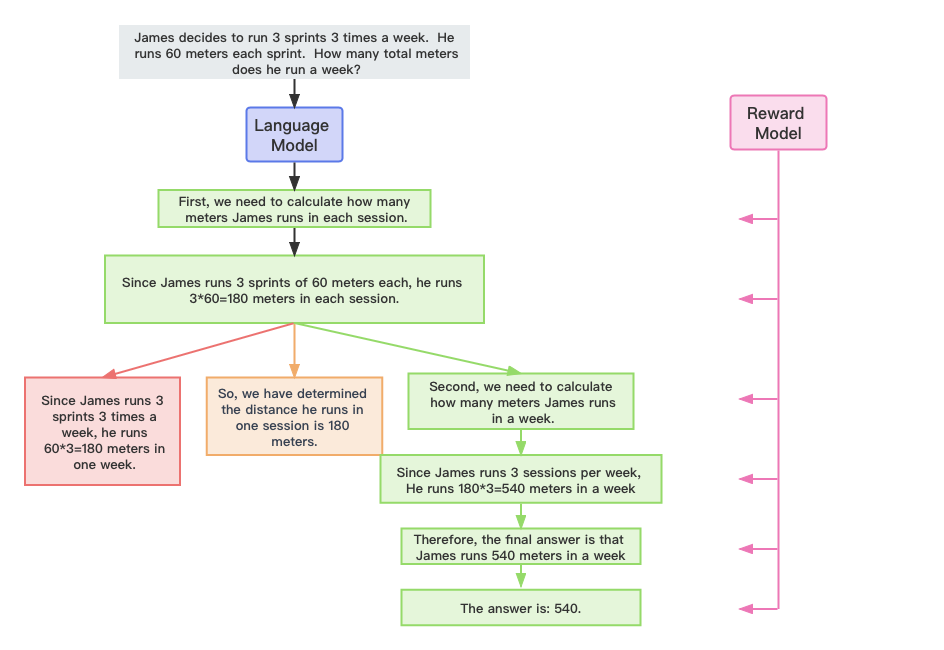}
    \caption{An example of HGS-PRM process.\colorbox{negative}{Red} denotes a negative step, \colorbox{positive}{Green} indicates a positive step, and \colorbox{neutral}{Orange} signifies a neutral step.}
    \label{fig: HGS-PRM example}
\end{figure}

Consequently, we propose a Heuristic Greedy search algorithm for large language models based on feedback from the process-supervised reward model(Figure \ref{fig: greedy search}) as detailed in Appendix~\ref{app: HGS-PRM appendix}. Figure \ref{fig: HGS-PRM example} shows an example of HGS-PRM for a GSM8K problem.

The Algorithm in the Appendix\ref{app: HGS-PRM appendix} illustrates the process of our heuristic greedy search algorithm. Given an input question \(x\), our algorithm expands to the next potential node and then evaluates it using a reward model. If the reward is positive, the algorithm advances using this new node. If not, it continues to explore the next potential node, assessing it with the reward model until it reaches the maximum bandwidth \(B\). If a node expands to the maximum bandwidth without finding a positive step, two scenarios are considered: If neutral sub-nodes exist, one is selected. If all sub-nodes are negative, this indicates an erroneous step. The large language model is unable to determine the correct subsequent step, prompting a backtrack. To constrain our search space efficiently and balance exploration with efficiency, we've set max bandwidth \(B\) and max iterations \(T\) parameters. Once the algorithm reaches the max iteration count, it will generate an answer from the current step. In our actual implementation, we also tried BFS and MCTS. However, we found that they introduced a more redundant search space and did not effectively improve the results. Our ultimate goal is to enhance the model's mathematical reasoning ability, not to increase algorithmic complexity. Therefore, the heuristic greedy search achieves a better trade-off between exploration and efficiency and is a more optimal choice.

\subsection{Process-supervised Data for Code}
To extend our approach to the field of code, a step-level reward dataset for code is necessary. As opposed to maths problems where it is difficult to assess the correctness of a given reasoning step without human intervention, coding problems with unit-tests are uniquely suited for automated labeling of step-level reward data. In our work, we use MBPP as seed dataset from which our PRM for code dataset is created.

\begin{figure}[H]
    \centering
    \includegraphics[width=0.75\linewidth]{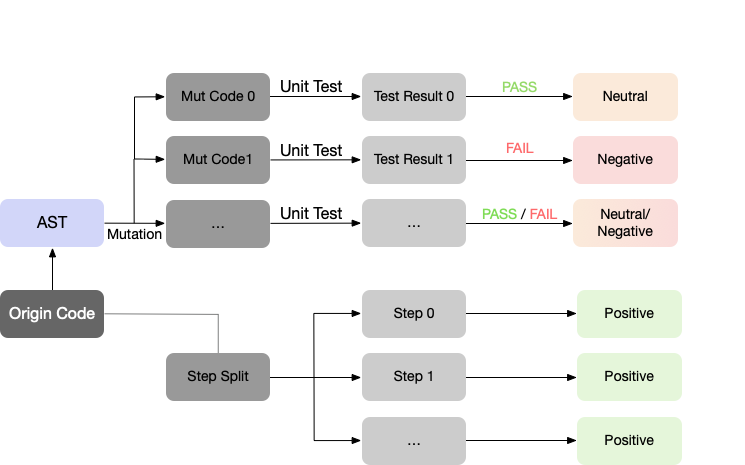}
    \caption{\textbf{The process of generating the PRM-Code dataset with ground truth code and unit test.}}
    \label{fig:code-data}
\end{figure}

Figure \ref{fig:code-data} provides an illustration of our method for generating code PRM data. Similar to PRM data where the validity of each reasoning step is evaluated in the context of previous steps, the format of our PRM-Code data code also includes components such as (\textbf{PREVIOUS DATA}, \textbf{STEP} and \textbf{LABEL}). As Python programs are naturally divided into lines separated by the new-line token, we first collect 
\colorbox{positive}{positive} results from the ground truth data.
For example, if we select the n-th step as the current annotated step, the \textbf{PREVIOUS DATA} includes the prompt and steps 0 to n-1, the \textbf{STEP} is the n-th step, and the \textbf{LABEL} is \colorbox{positive}{positive}.

The creation of step-level reward data with neutral and negative reward for code can the be accomplished with a combination of code mutation and unit-testing. Automatic code mutation (as in Mutation Testing) is a standard practice in code development where operators (such as addition) in a given code block is automatically mutated to similar operators (such as division). 
Such mutation requires first transforming a given code block into an abstract syntax tree (AST) representation, and defining rules of how different nodes on the AST are to be mutated. This yields N mutated code variants for each line of code where each mutated code is executable, and each variant undergoes only one mutation. These mutated codes are then executed with unit tests. If a mutated code passes the unit tests, its \textbf{LABEL} for the corresponding step is marked as \colorbox{neutral}{neutral}. If it fails to pass the unit tests, the \textbf{LABEL} is marked as \colorbox{negative}{negative}. The \textbf{PREVIOUS DATA} includes the prompt and steps 0 to the previous step of the mutation step, the STEP corresponds to the mutation step generated.

We note that mutation testing primarily focus on atomic operators such as arithmetic operations and array slicing, which limits the generality of our PRM-Code dataset. Natural extensions of our current approach involves defining more elaborate mutation rules that involve, for example, mutating for-loops into while-loops that correspond to changes of multiple lines of code. However, despite this limitation, our PRM-Code dataset has proven effectively in improving the code generation performance as we demonstrate in subsequent sections.

%% file: experiment.tex
\section{Experiment}
\label{headings}

In this section, we will delve into the intricacies of training our processed reward model and present the evaluation results of our heuristic greedy search algorithm, based on this processed reward model, in mathematical and code generation tasks.

\subsection{Process-supervised reward model training}
For mathematical tasks, we trained our process-supervised reward model(PRM) based on LLaMA-7B\citep{touvron2023llama}. As previously mentioned, our model training method first involved directive fine-tuning using the MATH training set, followed by reward model training. However, it should be noted that we also directly trained our reward model on LLaMA-7B. Our experimental results indicate that models fine-tuned with mathematical directives perform superiorly in all aspects compared to the base model.

\begin{figure}[h]
    \centering
    \includegraphics[width=0.7\linewidth]{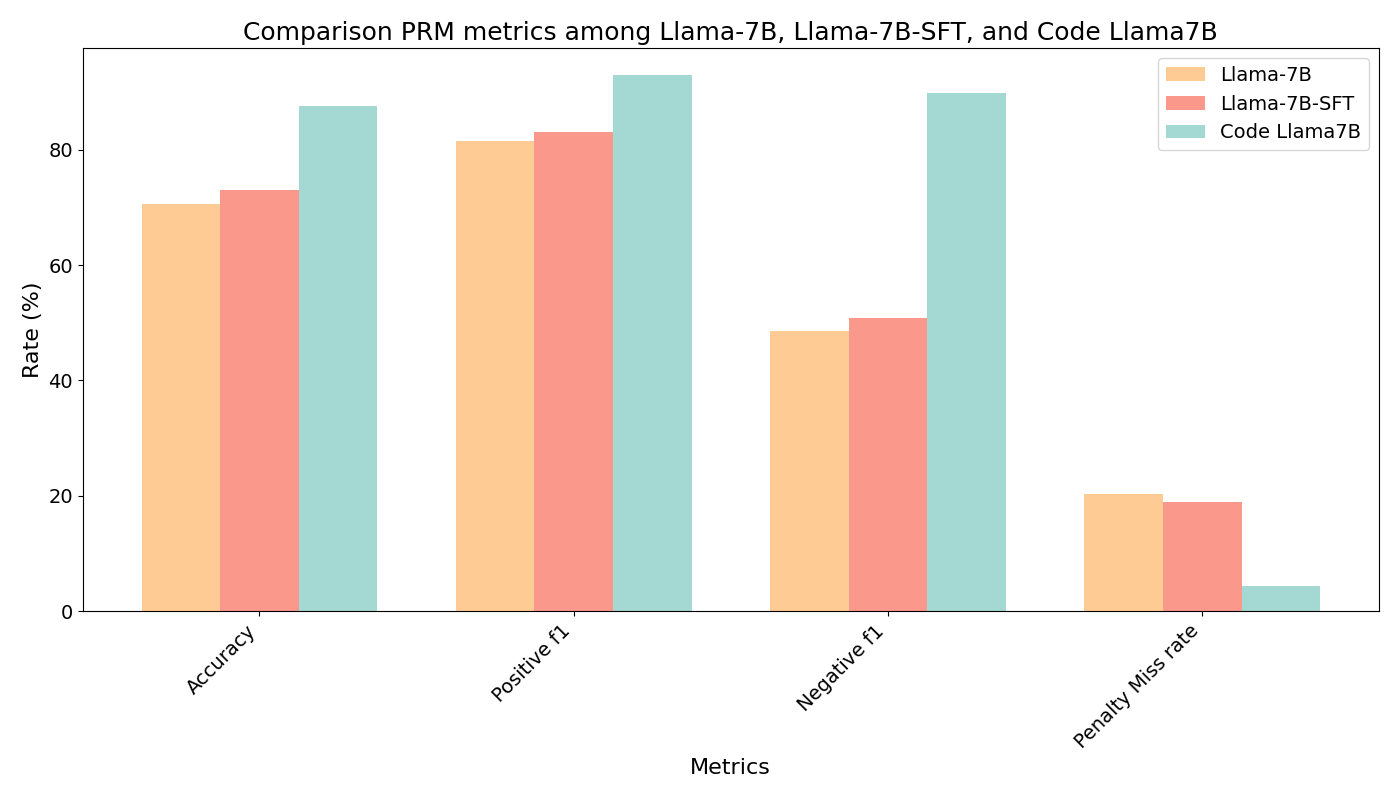}
    \caption{Key metrics of different processed-supervised reward model}
    \label{fig: PRM metrics}
\end{figure}

From our result shown in Figure \ref{fig: PRM metrics}, the Instruction-tuning model with math data before the training reward model results in improved performance. We also find positive labels exhibit the highest precision and recall. Therefore, we adopted a greedy strategy when designing our search algorithm, directly selecting the step when a positive step is identified. We also find The ability to distinguish neutral labels is inadequate, so give them less consideration during decoding. They can be treated as either positive or negative. 

Penalty miss rate refers to the proportion of classifying positives as negatives and negatives as positives in the mis-classification, which we consider to be the most serious classification error of the reward model. For neutral samples, We can avoid misjudgments by adjusting the action of our search algorithm for neutral samples.

We also trained a process-supervised reward model specifically for code. Drawing from our experiments with math, we directly use Code-LLaMA-7B\citep{rozière2023code}, which had been fine-tuned with code instructions, to train our process-supervised reward model for code. Our code step-level data was automatically generated on MBPP\citep{austin2021program} with the method in Figure \ref{fig:code-data}.

From our result in Figure \ref{fig: PRM metrics}, the process-supervised reward model(PRM) demonstrated a significantly better ability to discern negative cases in code compared to math. This suggests that there's greater potential in applying the PRM within the coding domain.

\subsection{Math result}

For mathematical problems, we first conducted evaluation experiments on the general models LLaMA-7B and LLaMA-13B using GSM8K and MATH datasets. We achieved results on both GSM8K and MATH datasets that were superior to those of CoT. For the MATH task, we follow the setting of \citep{lightman2023lets} and evaluate our models only on the 500 MATH test problems because 4.5K MATH test problems are in the PRM800K training set, which we used to train the reward model. Apart from general models, we also conducted experiments on math-specific models WizardMath-7B and WizardMath-13B\citep{luo2023wizardmath}. We similarly achieved superior results to CoT on both the GSM8K and MATH datasets. Moreover, we observed that math-specific models benefit more from the reward model. If the language model's intrinsic capability is too weak, even with the aid of a reward model, it remains challenging to sample the correct reasoning path. On the other hand, if the linguistic capacity of the model significantly surpasses that of the reward model, the benefits might not be pronounced. Therefore, aligning the capabilities of the reward model and the language model is of paramount importance.

\begin{table}[H]
\centering
\caption{We adopted the sampling setting with temperature=0.1 and top-p=0.95. For LLaMA, we use an 8-shot prompt for GS8K and a 4-shot prompt for the MATH task. For WizardMath, we follow the \cite{luo2023wizardmath} prompt}
\label{tab: math table}
\begin{tabular}{cccc}
\toprule
\textbf{model} & \textbf{method} & \textbf{GSM8K} & \textbf{MATH} \\
\midrule
LLaMA2-7B & CoT & 15.7\% & 2.4\% \\
LLaMA2-7B & HGS-PRM & 16.2\%(\textcolor{red}{+0.5\%}) & 2.6\%(\textcolor{red}{+0.2\%}) \\
LLaMA2-13B & CoT & 31.7\% & 2.8\% \\
LLaMA2-13B & HGS-PRM & 32.9\%(\textcolor{red}{+1.2\%}) & 4.1\%(\textcolor{red}{+1.3\%}) \\
WizardMath-7B & CoT & 54.3\% & 9.7\% \\
WizardMath-7B & HGS-PRM & 57.2\%(\textcolor{red}{+2.9\%}) & 11.1\%(\textcolor{red}{+1.4\%}) \\
WizardMath-13B & CoT & 63.2\% & 10.4\% \\
WizardMath-13B & HGS-PRM & 65.4\%(\textcolor{red}{+2.2\%}) & 13.7\%(\textcolor{red}{+3.3\%}) \\
\bottomrule
\end{tabular}
\end{table}

\subsection{Code result}

To avoid data leakage, our code PRM's step-level training data is created based on MBPP dataset\citep{austin2021program}, and the resulting reward model is evaluated on the HumanEval\citep{chen2021codex} benchmark. HumanEval is a widely recognized benchmark for testing large language models on coding tasks. It consists of 164 original programming questions, with an average of 9.6 test cases assigned to each question.

\begin{table}[H]
\centering
\caption{We follow the previous works \cite{chen2021codex} to estimate the pass@1 score with the same set of hyper-parameters: temperate=0.2, and top-p=0.95}
\label{tab: code table}
\begin{tabular}{ccc}
\toprule
\textbf{model} & \textbf{method} & \textbf{HumanEval pass@1} \\
\midrule
Code-LLaMA-Python-7B & CoT & 36.6\% \\
Code-LLaMA-Python-7B & HGS-PRM & 41.5\%(\textcolor{red}{+4.9\%}) \\
Code-LLaMA-Python-13B & CoT & 41.5\% \\
Code-LLaMA-Python-13B & HGS-PRM & 44.5\%(\textcolor{red}{+3.0\%}) \\
\bottomrule
\end{tabular}
\end{table}

For our evaluation, we selected models Code-LLaMA-Python-7B and Code-LLaMA-Python-13B. On the HumanEval benchmark, our approach improved pass@1 result by 4.9\%(36.6\% VS 41.5\%) for Code-LLaMA-Python-7B model and 3.0\% (41.5\% VS 44.5\%) for Code-LLaMA-Python-13B model over CoT, respectively. This performance also surpassed the greedy decoding results presented in the Code-LLaMA paper\citep{rozière2023code}, which is 3.1\%(38.4\% VS 41.5\%) and 1.2\%(43.3\% VS 44.5\%). The accuracy improvement we observed in code is generally higher than in mathematics. This aligns with our prior evaluation results during the PRM training reward phase where the precision and recall for the Code PRM negative label were notably higher than for math. We hypothesize that this might be because both HumanEval and MBPP involve relatively simple programming challenges, whereas MATH presents more complex mathematical problems which are intrinsically more challenging for both PRM and the language models themselves to learn.

\subsection{Further Analysis}

Is the result obtained through HGS-PRM reliable? We further analyzed the relationship between PRM feedback and the correctness of the results. The actual process is often not so ideal. As indicated in \ref{fig: PRM metrics}, the actual precision of PRM is lower than 90\%. If the accuracy of PRM is not up to par, HGS-PRM might reduce the accuracy of the results. Therefore, besides the greedy search method, we can also use the approach of scoring after sampling to verify whether PRM is effective. 

For the field of mathematics, we sampled 10 times from the MATH dataset with LLaMA2-13B and then scored the sampled reasoning paths at the step-level using PRM. According to our HGS-PRM algorithm, the correct reasoning path should not have any negative steps, so we filtering all samples with negative steps. For code, we sampled 100 times from the HumanEval dataset using Star-Coder\citep{li2023starcoder} and scored in the same manner.

From our results in Table \ref{tab: math sample table}, the ``Accuracy after filtering''(14.4\%, 36.66\%) is much higher than the original accuracy(4.25\%, 30.13\%), demonstrating the correlation between the step-level feedback provided by PRM and the correctness of the result. The computational cost of filtering after repeated sampling is higher than that of HGS-PRM, because, for complex reasoning problems, it's challenging for large language models to sample reasoning paths that don't contain negative steps. This is also why we use search instead of sampling.

\begin{table}[H]
\centering
\caption{We adopted the sampling setting with temperature=0.8 and top-p=0.95 for reasoning path sampling using LLaMA2-13B model. The \textbf{Accuracy after filtering} refers to the accuracy of sampled responses filtered by PRM model that contain no negative reasoning steps.}
\label{tab: math sample table}
\begin{tabular}{ccc}
\toprule
\textbf & \textbf{MATH} & \textbf{HumanEval} \\
\midrule
\textbf{Accuracy} & 4.25\% & 30.13\%\\
\textbf{Accuracy after filtering} & 14.4\% & 36.66\%\\
\bottomrule
\end{tabular}
\end{table}

\subsection{Case Study}
Appendix~\ref{app: Case study appendix} displays some correct reasoning paths and erroneous paths obtained by HGS-PRM, further detailing what mistakes PRM specifically identified such as parsing errors, computational errors, and semantic errors.

%% file: related.tex
\section{related Works}

\label{others}

\textbf{Large Language Models For reasoning}
Large language models (LLMs) have faced challenges in complex reasoning tasks, covering mathematical reasoning \citep{lu2023survey, frieder2023mathematical} and code generation \citep{chen2021codex,li2023starcoder,rozière2023code}. Key datasets for mathematical reasoning are GSM8K \citep{cobbe2021training}, MATH \citep{hendrycksmath2021}, AddSub \citep{hosseini-etal-2014-learning}, MultiArith \citep{roy-roth-2015-solving}, and SingleEQ \citep{koncel-kedziorski-etal-2015-parsing} while code generation evaluations employ HumanEval \citep{chen2021codex}, MBPP \citep{austin2021program}, and DS1000 \citep{Lai2022DS1000}.. \cite{wei2023chainofthought} proposed "Chain-of-Thought Prompting" (CoT) to improve LLMs' performance by prompting step-by-step decompositions. The Zero-shot-CoT by \cite{kojima2023large} enhances this by adding 'Let’s think step by step' to responses, detailing the model's reasoning. \cite{gao2023pal} introduced the Program-aided Language Model (PAL) to assist LLMs in reasoning, while \cite{wang2023selfconsistency} presented a self-consistent method building upon CoT for consistent responses. \cite{zhao2023automatic} combined both CoT and PAL for improved reasoning, allowing concurrent use with the self-consistent approach. Lastly, \cite{zhang2022automatic}'s Auto-CoT leverages LLMs to produce diverse reasoning chains autonomously.

\textbf{Reasoning with feedback}
Beyond prompt-based enhancements, \cite{pan2023automatically} reviews methods combining self-feedback, search, and tools \citep{gou2023critic} to amplify large language model reasoning. Research by \cite{madaan2023selfrefine}, \cite{gero2023selfverification}, \cite{chen2023teaching}, \cite{weng2023large} and \cite{shinn2023reflexion} demonstrate that these models can serve as feedback sources to bolster reasoning. \cite{xie2023decomposition} offers a prompting method integrating self-evaluation via stochastic beam search to refine multi-step inference. \cite{yao2023tree}'s "ToT" (Tree of Thoughts) advances CoT, enabling models to explore varied reasoning paths, evaluate decisions, and adjust strategies. \cite{hao2023reasoning} transforms the LLM to act as both a world model and logical agent, using a planning method inspired by Monto Carlo Tree Search. \cite{wang2023leti}'s LeTI uses textual feedback from code errors for better code generation. \cite{yang2023leandojo} takes a retrieval approach, allowing LLMs to extract data from Lean and engage with proof environments. Finally, \cite{yao2023react} presents a framework blending advances in reasoning and action for a broader spectrum of linguistic challenges.

\textbf{PRM \& ORM}

While much attention has been given to other areas, the process-supervised reward model (PRM) and outcome-supervised reward model (ORM) have seen less exploration. \cite{uesato2022solving} first introduced PRM, highlighting its advantages over ORM in several applications, from few-shot prompting to reward modeling. Expanding on this, \cite{lightman2023lets} released PRM800K, a dataset based on MATH annotations, showcasing the reliability of process supervision over outcome supervision. This high-quality dataset has been invaluable to our research. \cite{luo2023wizardmath} introduced "Reinforcement Learning from Evol-Instruct Feedback (RLEIF)", using PRM as a reward model within the PPO framework \citep{schulman2017proximal}. While these studies have focused on PRM for math, there's a noticeable gap in PRM research for coding, pointing to a ripe area for further investigation.

%% file: conclusion.tex
\section{Conclusions and Future Work}
In this study, we introduce an approach that integrates the process-supervised reward Model (PRM) into heuristic greedy search, demonstrating the value of incorporating PRM feedback in multi-step inference tasks within large language models. This offers fresh insights into utilizing the PRM framework rather than RLHF or reject sampling. Our experimental results on the MATH and GSM8K datasets validate the effectiveness of our proposed HGS-PRM method and explore the potential of PRM in the decoding phase experiment.

Moreover, we take advantage of the Mutating Testing technique for code PRM data generation, enabling rapid scaling of step-level reward data for code applications. We release a PRM dataset tailored for coding, which holds pivotal significance for further research on PRM's application in code-related tasks. We believe that the potential of the process-supervised reward model in multi-step reasoning within expansive models remains under-explored, especially in coding tasks. Enhancing code-specific PRM data and reward models could significantly bolster the capabilities of large language models in code generation.



%% file: appendix.tex
\section{Heuristic Greedy Search with PRM(HGS-PRM)}
\label{app: HGS-PRM appendix}

\begin{enumerate}
\item \textbf{Node definition} Each node of the search tree possesses the following elements.
    \begin{enumerate}
        \item \textbf{State} \( S = [x, s_1 ... s_i]\), where the input prompt and question \(Q\) represented as \(x\), each step \(i\) represented as \(s_i\)
        \item \textbf{Step} \( s_i\), the reasoning step for this node
        \item \textbf{Reward} \( r\), the reward for the step \(i\)(\( s_i\)) evaluated from the PRM
        \item \textbf{Value} \( V\), the cumulative reward for each step leading to that node.
        \item \textbf{Parent} \( parent\), the parent of the node
        \item \textbf{Children} \( children\), the children node of the node
    \end{enumerate}
\item \textbf{Expand} Expand the possible next steps from the current node, and create a new child node:\(P(s_{i+1}|S) = P(s_{i+1}|x, s_1···s_i)\). the sampling diversity is influenced by the sample hyper-parameter like temperature, top-\(p\), top-\(k\).
\item \textbf{Get Reward}: after expanding a new node, the process reward model will evaluate the new step and feedback reward range from \([-1,0,1]\), which means \colorbox{negative}{negative}, \colorbox{neutral}{neutral}, \colorbox{positive}{positive} step.
\item \textbf{Go ahead}: when the reward of the step is \colorbox{positive}{positive}(r = 1), we will choose this step and go ahead to this new child node.
\item \textbf{Backup}: When this node cannot expand into different steps or the number of child nodes reaches the maximum bandwidth, it will backtrack to its parent node.
\end{enumerate}

\begin{algorithm}[H]
\caption{Heuristic Greedy Search with PRM(HGS-PRM)}
\label{alg: greedy search}
\begin{algorithmic}[1]
\REQUIRE Input Question \(x\), LLM generate function\(P()\), next step Expand function E(), get reward function \(R()\), max bandwidth \(B\), max iterations \(T\), end token or end sequence of LLM \(eos\_ token\)
\STATE $cur\_node \leftarrow \{x\}$
\FOR{$i = 1$ to $T$}
    \IF{number of $cur\_node.child < B$}
        \STATE $next\_node \leftarrow E(P, cur\_node)$
        \IF{$eos\_token$ in $next\_node.step$}
            \RETURN $next\_node.state$
        \ELSIF{$next\_node.reward == 1$}
            \STATE $cur\_node \leftarrow next\_node$
        \ENDIF
    \ELSE
        \IF{$child.reward = 0 $ for any child in $cur\_node.children$}
            \STATE $cur\_node \leftarrow $ one of the child
        \ELSE
            \STATE $cur\_node \leftarrow cur\_node.parent$
        \ENDIF
    \ENDIF
\ENDFOR
\STATE $result \leftarrow P(cur\_node.state)$
\RETURN $result$
\end{algorithmic}
\end{algorithm}

\section{Case study}
\label{app: Case study appendix}
\subsection{MATH Case study}

\begin{table}[H]
\centering
\label{tab: wizard math example}
\begin{tabular}{p{3cm}|p{10cm}}
\toprule
\textbf{Question} & Riproarin' Ringo was roping a recalcitrant dogie. Ringo decided to give the dogie a reprieve by calculating \[|(1-i)^8|\]before riding after the dogie. What answer should Ringo have found? \\
\midrule
\textbf{Ground Truth} & \textbf{540} \\
\midrule
\textbf{Correct path} &  \colorbox{positive}{\parbox{10cm}{We are given a complex number in the form of $(1-i)$, and we need to find the magnitude of its eightfold, which is $(1-i)^8$.}}
\colorbox{positive}{\parbox{10cm}{The magnitude of a product is the product of the magnitudes, i.e., $|ab| = |a||b|$. In our case, we have $(1-i)^8$, and we want to find $|(1-i)^8|$.}}
\colorbox{positive}{\parbox{10cm}{Using the property from Step 2, we can rewrite $|(1-i)^8|$ as $|1||(1-i)|^8|1|$. Since the magnitude of a number is always non-zero, we can safely cancel out the $|1|$ terms, leaving us with $|(1-i)|^8$.}}
\colorbox{positive}{\parbox{10cm}{Now, we need to find the magnitude of $(1-i)$. The magnitude of a complex number $a+bi$ is given by $\sqrt{a^2 + b^2}$. In our case, $a = 1$ and $b = -1$, so the magnitude of $(1-i)$ is $\sqrt{1^2 + (-1)^2} = \sqrt{2}$.}}
\colorbox{positive}{\parbox{10cm}{Finally, we need to find $|(1-i)^8|$. Since we found that $|(1-i)| = \sqrt{2}$, we can raise it to the power of 8: $(\sqrt{2})^8$.}}
\colorbox{positive}{\parbox{10cm}{We can simplify $(\sqrt{2})^8$ as $2^{\frac{8}{2}}$. Since $\frac{8}{2}$ is equal to 4, we have $2^4 = 16$.}}
\colorbox{positive}{\parbox{10cm}{So, the final answer is $\boxed{16}$.}}
\colorbox{positive}{\parbox{10cm}{The answer is: 16.}}\\
\midrule
\parbox{3cm}{\textbf{Wrong path}}
\parbox{3cm}{(computational errors)}
& \colorbox{positive}{\parbox{10cm}{We are given a complex number in the form of $(1-i)$, and we need to find the magnitude of its eightfold, which is $(1-i)^8$.}}
\colorbox{positive}{\parbox{10cm}{Before finding the magnitude of $(1-i)^8$, we need to find the magnitude of $(1-i)^2$. To do this, we can use the formula for the square of a complex number: $(a+bi)^2 = a^2 + 2abi + b^2$. In our case, $a = 1$ and $b = -1$, so $(1-i)^2 = (1)^2 + 2(1)(-1) + (-1)^2 = 1 - 2i + 1 = 2 - 2i$.}}
\colorbox{neutral}{\parbox{10cm}{Now, we need to find the magnitude of $(2-2i)^4$. Using the same formula for the square of a complex number, we get $(2-2i)^4 = (2)^4 + 2(2)(-2)i + (-2)^4 = 16 - 32i + 16 = 32 - 32i$.}}
\colorbox{negative}{\parbox{10cm}{Next, we need to find the magnitude of $(32-32i)^2$. Using the formula for the square of a complex number again, we get $(32-32i)^2 = (32)^2 + 2(32)(-32)i + (-32)^2 = 1024 - 2048i + 1024 = 2048 - 2048i$.}}
Finally, we need to find the magnitude of $(2048-2048i)^4$. Using the formula for the square of a complex number one more time, we get $(2048-2048i)^4 = (2048)^4 + 2(2048)(-2048)i + (-2048)^4 = 16,777,216 - 4,195,392i + 16,777,216 = 33,554,508$.
Now, we take the absolute value of the result: $|33,554,508| = 33,554,508$.
So, the final answer is $\boxed{33,554,508}$.
The answer is: 33,\!554,\!508.\\
\bottomrule
\end{tabular}
\caption{MATH case with WizardMath-13B}
\end{table}

\subsection{GSM8K Case study}

\begin{table}[H]
\centering
\label{tab: wizard math example}
\begin{tabular}{p{3cm}|p{10cm}}
\toprule
\textbf{Question} & James decides to run 3 sprints 3 times a week.  He runs 60 meters each sprint.  How many total meters does he run a week? \\
\midrule
\textbf{Ground Truth} & \textbf{540} \\
\midrule
\textbf{Correct path} &  \colorbox{positive}{\parbox{10cm}{First, we need to calculate how many meters James runs in each session.}}
\colorbox{positive}{\parbox{10cm}{Since he runs 3 sprints of 60 meters each, we can calculate the total distance he runs in each session by multiplying the number of sprints by the distance of each sprint.}}
\colorbox{positive}{\parbox{10cm}{Therefore, James runs 3*60=180 meters in each session.}}
\colorbox{positive}{\parbox{10cm}{Second, we need to calculate how many meters James runs in a week.}}
\colorbox{positive}{\parbox{10cm}{Since he runs 3 sessions per week, we can calculate the total distance he runs in a week by multiplying the distance he runs in each session by the number of sessions he runs per week.}}
\colorbox{positive}{\parbox{10cm}{Therefore, James runs 180*3=540 meters in a week.}}
\colorbox{positive}{\parbox{10cm}{Therefore, the final answer is that James runs 540 meters in a week.}}
\colorbox{positive}{\parbox{10cm}{The answer is: 540.}}\\
\midrule
\parbox{3cm}{\textbf{Wrong path}}
\parbox{3cm}{(parsing errors)}
& \colorbox{positive}{\parbox{10cm}{First, we need to calculate how many meters James runs in one sprint.}}
\colorbox{positive}{\parbox{10cm}{According to the question, James runs 60 meters each sprint. }}
\colorbox{neutral}{\parbox{10cm}{Therefore, we can calculate the total distance he runs in one week by multiplying the distance of one sprint by the number of sprints he runs in one week. }}
\colorbox{negative}{\parbox{10cm}{Since James runs 3 sprints 3 times a week, he runs 60*3=180 meters in one week.}}
Therefore, the total distance James runs in one week is 180 meters. 
Thus, the final answer is 180.
The answer is: 180. \\
\bottomrule
\end{tabular}
\caption{GSM8K case with WizardMath-13B}
\end{table}

\newpage
\subsection{HumanEval case study}

\lstset{
  language=Python,
  basicstyle=\ttfamily\small,
  escapeinside={(*@}{@*)}
}

\begin{table}[h]
\centering
\begin{tabular}{p{3cm}|p{12cm}}
\hline
\textbf{Prompt} &
\begin{lstlisting}
def how_many_times(string: str, substring: str) -> int:
    """ Find how many times a given substring 
    can be found in the original string.
    Count overlaping cases.
    >>> how_many_times('', 'a')
    0
    >>> how_many_times('aaa', 'a')
    3
    >>> how_many_times('aaaa', 'aa')
    3
    """
\end{lstlisting} \\
\hline
\textbf{Correct path} & 
\begin{lstlisting}
    (*@\colorbox{positive}{if len(substring) == 0:}@*)
        (*@\colorbox{positive}{return 0}@*)
    (*@\colorbox{positive}{if len(string) == 0:}@*)
        (*@\colorbox{positive}{return 0}@*)
    (*@\colorbox{positive}{if len(string) < len(substring):}@*)
        (*@\colorbox{positive}{return 0}@*)
    (*@\colorbox{positive}{count = 0}@*)
    (*@\colorbox{positive}{for i in range(len(string) - len(substring) + 1):}@*)
        (*@\colorbox{positive}{if string[i:i + len(substring)] == substring:}@*)
            (*@\colorbox{positive}{count += 1}@*)
    (*@\colorbox{positive}{return count}@*)
\end{lstlisting} \\
\hline
\parbox{3cm}{\textbf{Wrong path}}
\parbox{3cm}{(semantic errors)}
& 
\begin{lstlisting}
    (*@\colorbox{neutral}{if len(substring) == 0:}@*)
        (*@\colorbox{positive}{return 0}@*)
    (*@\colorbox{negative}{return string.replace(substring, '', 1).count(substring)}@*)
\end{lstlisting} \\
\hline
\end{tabular}
\caption{HumanEval case with Code-LLaMA-Python-13B}
\end{table}

%% file: iclr2023_conference.bbl
\begin{thebibliography}{51}
\providecommand{\natexlab}[1]{#1}
\providecommand{\url}[1]{\texttt{#1}}
\expandafter\ifx\csname urlstyle\endcsname\relax
  \providecommand{\doi}[1]{doi: #1}\else
  \providecommand{\doi}{doi: \begingroup \urlstyle{rm}\Url}\fi

\bibitem[Anil et~al.(2023)Anil, Dai, Firat, Johnson, Lepikhin, Passos, Shakeri, Taropa, Bailey, Chen, Chu, Clark, Shafey, Huang, Meier-Hellstern, Mishra, Moreira, Omernick, Robinson, Ruder, Tay, Xiao, Xu, Zhang, Abrego, Ahn, Austin, Barham, Botha, Bradbury, Brahma, Brooks, Catasta, Cheng, Cherry, Choquette-Choo, Chowdhery, Crepy, Dave, Dehghani, Dev, Devlin, Díaz, Du, Dyer, Feinberg, Feng, Fienber, Freitag, Garcia, Gehrmann, Gonzalez, Gur-Ari, Hand, Hashemi, Hou, Howland, Hu, Hui, Hurwitz, Isard, Ittycheriah, Jagielski, Jia, Kenealy, Krikun, Kudugunta, Lan, Lee, Lee, Li, Li, Li, Li, Li, Lim, Lin, Liu, Liu, Maggioni, Mahendru, Maynez, Misra, Moussalem, Nado, Nham, Ni, Nystrom, Parrish, Pellat, Polacek, Polozov, Pope, Qiao, Reif, Richter, Riley, Ros, Roy, Saeta, Samuel, Shelby, Slone, Smilkov, So, Sohn, Tokumine, Valter, Vasudevan, Vodrahalli, Wang, Wang, Wang, Wang, Wieting, Wu, Xu, Xu, Xue, Yin, Yu, Zhang, Zheng, Zheng, Zhou, Zhou, Petrov, and Wu]{anil2023palm}
Rohan Anil, Andrew~M. Dai, Orhan Firat, Melvin Johnson, Dmitry Lepikhin, Alexandre Passos, Siamak Shakeri, Emanuel Taropa, Paige Bailey, Zhifeng Chen, Eric Chu, Jonathan~H. Clark, Laurent~El Shafey, Yanping Huang, Kathy Meier-Hellstern, Gaurav Mishra, Erica Moreira, Mark Omernick, Kevin Robinson, Sebastian Ruder, Yi~Tay, Kefan Xiao, Yuanzhong Xu, Yujing Zhang, Gustavo~Hernandez Abrego, Junwhan Ahn, Jacob Austin, Paul Barham, Jan Botha, James Bradbury, Siddhartha Brahma, Kevin Brooks, Michele Catasta, Yong Cheng, Colin Cherry, Christopher~A. Choquette-Choo, Aakanksha Chowdhery, Clément Crepy, Shachi Dave, Mostafa Dehghani, Sunipa Dev, Jacob Devlin, Mark Díaz, Nan Du, Ethan Dyer, Vlad Feinberg, Fangxiaoyu Feng, Vlad Fienber, Markus Freitag, Xavier Garcia, Sebastian Gehrmann, Lucas Gonzalez, Guy Gur-Ari, Steven Hand, Hadi Hashemi, Le~Hou, Joshua Howland, Andrea Hu, Jeffrey Hui, Jeremy Hurwitz, Michael Isard, Abe Ittycheriah, Matthew Jagielski, Wenhao Jia, Kathleen Kenealy, Maxim Krikun, Sneha Kudugunta, Chang
  Lan, Katherine Lee, Benjamin Lee, Eric Li, Music Li, Wei Li, YaGuang Li, Jian Li, Hyeontaek Lim, Hanzhao Lin, Zhongtao Liu, Frederick Liu, Marcello Maggioni, Aroma Mahendru, Joshua Maynez, Vedant Misra, Maysam Moussalem, Zachary Nado, John Nham, Eric Ni, Andrew Nystrom, Alicia Parrish, Marie Pellat, Martin Polacek, Alex Polozov, Reiner Pope, Siyuan Qiao, Emily Reif, Bryan Richter, Parker Riley, Alex~Castro Ros, Aurko Roy, Brennan Saeta, Rajkumar Samuel, Renee Shelby, Ambrose Slone, Daniel Smilkov, David~R. So, Daniel Sohn, Simon Tokumine, Dasha Valter, Vijay Vasudevan, Kiran Vodrahalli, Xuezhi Wang, Pidong Wang, Zirui Wang, Tao Wang, John Wieting, Yuhuai Wu, Kelvin Xu, Yunhan Xu, Linting Xue, Pengcheng Yin, Jiahui Yu, Qiao Zhang, Steven Zheng, Ce~Zheng, Weikang Zhou, Denny Zhou, Slav Petrov, and Yonghui Wu.
\newblock Palm 2 technical report, 2023.

\bibitem[Austin et~al.(2021)Austin, Odena, Nye, Bosma, Michalewski, Dohan, Jiang, Cai, Terry, Le, and Sutton]{austin2021program}
Jacob Austin, Augustus Odena, Maxwell Nye, Maarten Bosma, Henryk Michalewski, David Dohan, Ellen Jiang, Carrie Cai, Michael Terry, Quoc Le, and Charles Sutton.
\newblock Program synthesis with large language models, 2021.

\bibitem[Bai et~al.(2022)Bai, Kadavath, Kundu, Askell, Kernion, Jones, Chen, Goldie, Mirhoseini, McKinnon, Chen, Olsson, Olah, Hernandez, Drain, Ganguli, Li, Tran-Johnson, Perez, Kerr, Mueller, Ladish, Landau, Ndousse, Lukosuite, Lovitt, Sellitto, Elhage, Schiefer, Mercado, DasSarma, Lasenby, Larson, Ringer, Johnston, Kravec, Showk, Fort, Lanham, Telleen-Lawton, Conerly, Henighan, Hume, Bowman, Hatfield-Dodds, Mann, Amodei, Joseph, McCandlish, Brown, and Kaplan]{bai2022constitutional}
Yuntao Bai, Saurav Kadavath, Sandipan Kundu, Amanda Askell, Jackson Kernion, Andy Jones, Anna Chen, Anna Goldie, Azalia Mirhoseini, Cameron McKinnon, Carol Chen, Catherine Olsson, Christopher Olah, Danny Hernandez, Dawn Drain, Deep Ganguli, Dustin Li, Eli Tran-Johnson, Ethan Perez, Jamie Kerr, Jared Mueller, Jeffrey Ladish, Joshua Landau, Kamal Ndousse, Kamile Lukosuite, Liane Lovitt, Michael Sellitto, Nelson Elhage, Nicholas Schiefer, Noemi Mercado, Nova DasSarma, Robert Lasenby, Robin Larson, Sam Ringer, Scott Johnston, Shauna Kravec, Sheer~El Showk, Stanislav Fort, Tamera Lanham, Timothy Telleen-Lawton, Tom Conerly, Tom Henighan, Tristan Hume, Samuel~R. Bowman, Zac Hatfield-Dodds, Ben Mann, Dario Amodei, Nicholas Joseph, Sam McCandlish, Tom Brown, and Jared Kaplan.
\newblock Constitutional ai: Harmlessness from ai feedback, 2022.

\bibitem[Brown et~al.(2020)Brown, Mann, Ryder, Subbiah, Kaplan, Dhariwal, Neelakantan, Shyam, Sastry, Askell, Agarwal, Herbert-Voss, Krueger, Henighan, Child, Ramesh, Ziegler, Wu, Winter, Hesse, Chen, Sigler, Litwin, Gray, Chess, Clark, Berner, McCandlish, Radford, Sutskever, and Amodei]{brown2020language}
Tom~B. Brown, Benjamin Mann, Nick Ryder, Melanie Subbiah, Jared Kaplan, Prafulla Dhariwal, Arvind Neelakantan, Pranav Shyam, Girish Sastry, Amanda Askell, Sandhini Agarwal, Ariel Herbert-Voss, Gretchen Krueger, Tom Henighan, Rewon Child, Aditya Ramesh, Daniel~M. Ziegler, Jeffrey Wu, Clemens Winter, Christopher Hesse, Mark Chen, Eric Sigler, Mateusz Litwin, Scott Gray, Benjamin Chess, Jack Clark, Christopher Berner, Sam McCandlish, Alec Radford, Ilya Sutskever, and Dario Amodei.
\newblock Language models are few-shot learners, 2020.

\bibitem[Chen et~al.(2021)Chen, Tworek, Jun, Yuan, de~Oliveira~Pinto, Kaplan, Edwards, Burda, Joseph, Brockman, Ray, Puri, Krueger, Petrov, Khlaaf, Sastry, Mishkin, Chan, Gray, Ryder, Pavlov, Power, Kaiser, Bavarian, Winter, Tillet, Such, Cummings, Plappert, Chantzis, Barnes, Herbert-Voss, Guss, Nichol, Paino, Tezak, Tang, Babuschkin, Balaji, Jain, Saunders, Hesse, Carr, Leike, Achiam, Misra, Morikawa, Radford, Knight, Brundage, Murati, Mayer, Welinder, McGrew, Amodei, McCandlish, Sutskever, and Zaremba]{chen2021codex}
Mark Chen, Jerry Tworek, Heewoo Jun, Qiming Yuan, Henrique~Ponde de~Oliveira~Pinto, Jared Kaplan, Harri Edwards, Yuri Burda, Nicholas Joseph, Greg Brockman, Alex Ray, Raul Puri, Gretchen Krueger, Michael Petrov, Heidy Khlaaf, Girish Sastry, Pamela Mishkin, Brooke Chan, Scott Gray, Nick Ryder, Mikhail Pavlov, Alethea Power, Lukasz Kaiser, Mohammad Bavarian, Clemens Winter, Philippe Tillet, Felipe~Petroski Such, Dave Cummings, Matthias Plappert, Fotios Chantzis, Elizabeth Barnes, Ariel Herbert-Voss, William~Hebgen Guss, Alex Nichol, Alex Paino, Nikolas Tezak, Jie Tang, Igor Babuschkin, Suchir Balaji, Shantanu Jain, William Saunders, Christopher Hesse, Andrew~N. Carr, Jan Leike, Josh Achiam, Vedant Misra, Evan Morikawa, Alec Radford, Matthew Knight, Miles Brundage, Mira Murati, Katie Mayer, Peter Welinder, Bob McGrew, Dario Amodei, Sam McCandlish, Ilya Sutskever, and Wojciech Zaremba.
\newblock Evaluating large language models trained on code.
\newblock 2021.

\bibitem[Chen et~al.(2023)Chen, Lin, Schärli, and Zhou]{chen2023teaching}
Xinyun Chen, Maxwell Lin, Nathanael Schärli, and Denny Zhou.
\newblock Teaching large language models to self-debug, 2023.

\bibitem[Chowdhery et~al.(2022)Chowdhery, Narang, Devlin, Bosma, Mishra, Roberts, Barham, Chung, Sutton, Gehrmann, Schuh, Shi, Tsvyashchenko, Maynez, Rao, Barnes, Tay, Shazeer, Prabhakaran, Reif, Du, Hutchinson, Pope, Bradbury, Austin, Isard, Gur-Ari, Yin, Duke, Levskaya, Ghemawat, Dev, Michalewski, Garcia, Misra, Robinson, Fedus, Zhou, Ippolito, Luan, Lim, Zoph, Spiridonov, Sepassi, Dohan, Agrawal, Omernick, Dai, Pillai, Pellat, Lewkowycz, Moreira, Child, Polozov, Lee, Zhou, Wang, Saeta, Diaz, Firat, Catasta, Wei, Meier-Hellstern, Eck, Dean, Petrov, and Fiedel]{chowdhery2022palm}
Aakanksha Chowdhery, Sharan Narang, Jacob Devlin, Maarten Bosma, Gaurav Mishra, Adam Roberts, Paul Barham, Hyung~Won Chung, Charles Sutton, Sebastian Gehrmann, Parker Schuh, Kensen Shi, Sasha Tsvyashchenko, Joshua Maynez, Abhishek Rao, Parker Barnes, Yi~Tay, Noam Shazeer, Vinodkumar Prabhakaran, Emily Reif, Nan Du, Ben Hutchinson, Reiner Pope, James Bradbury, Jacob Austin, Michael Isard, Guy Gur-Ari, Pengcheng Yin, Toju Duke, Anselm Levskaya, Sanjay Ghemawat, Sunipa Dev, Henryk Michalewski, Xavier Garcia, Vedant Misra, Kevin Robinson, Liam Fedus, Denny Zhou, Daphne Ippolito, David Luan, Hyeontaek Lim, Barret Zoph, Alexander Spiridonov, Ryan Sepassi, David Dohan, Shivani Agrawal, Mark Omernick, Andrew~M. Dai, Thanumalayan~Sankaranarayana Pillai, Marie Pellat, Aitor Lewkowycz, Erica Moreira, Rewon Child, Oleksandr Polozov, Katherine Lee, Zongwei Zhou, Xuezhi Wang, Brennan Saeta, Mark Diaz, Orhan Firat, Michele Catasta, Jason Wei, Kathy Meier-Hellstern, Douglas Eck, Jeff Dean, Slav Petrov, and Noah Fiedel.
\newblock Palm: Scaling language modeling with pathways, 2022.

\bibitem[Cobbe et~al.(2021)Cobbe, Kosaraju, Bavarian, Chen, Jun, Kaiser, Plappert, Tworek, Hilton, Nakano, Hesse, and Schulman]{cobbe2021training}
Karl Cobbe, Vineet Kosaraju, Mohammad Bavarian, Mark Chen, Heewoo Jun, Lukasz Kaiser, Matthias Plappert, Jerry Tworek, Jacob Hilton, Reiichiro Nakano, Christopher Hesse, and John Schulman.
\newblock Training verifiers to solve math word problems, 2021.

\bibitem[Dong et~al.(2023)Dong, Xiong, Goyal, Zhang, Chow, Pan, Diao, Zhang, Shum, and Zhang]{dong2023raft}
Hanze Dong, Wei Xiong, Deepanshu Goyal, Yihan Zhang, Winnie Chow, Rui Pan, Shizhe Diao, Jipeng Zhang, Kashun Shum, and Tong Zhang.
\newblock Raft: Reward ranked finetuning for generative foundation model alignment, 2023.

\bibitem[Frieder et~al.(2023)Frieder, Pinchetti, Chevalier, Griffiths, Salvatori, Lukasiewicz, Petersen, and Berner]{frieder2023mathematical}
Simon Frieder, Luca Pinchetti, Alexis Chevalier, Ryan-Rhys Griffiths, Tommaso Salvatori, Thomas Lukasiewicz, Philipp~Christian Petersen, and Julius Berner.
\newblock Mathematical capabilities of chatgpt, 2023.

\bibitem[Gao et~al.(2023)Gao, Madaan, Zhou, Alon, Liu, Yang, Callan, and Neubig]{gao2023pal}
Luyu Gao, Aman Madaan, Shuyan Zhou, Uri Alon, Pengfei Liu, Yiming Yang, Jamie Callan, and Graham Neubig.
\newblock Pal: Program-aided language models, 2023.

\bibitem[Gero et~al.(2023)Gero, Singh, Cheng, Naumann, Galley, Gao, and Poon]{gero2023selfverification}
Zelalem Gero, Chandan Singh, Hao Cheng, Tristan Naumann, Michel Galley, Jianfeng Gao, and Hoifung Poon.
\newblock Self-verification improves few-shot clinical information extraction, 2023.

\bibitem[Gou et~al.(2023)Gou, Shao, Gong, Shen, Yang, Duan, and Chen]{gou2023critic}
Zhibin Gou, Zhihong Shao, Yeyun Gong, Yelong Shen, Yujiu Yang, Nan Duan, and Weizhu Chen.
\newblock Critic: Large language models can self-correct with tool-interactive critiquing, 2023.

\bibitem[Hao et~al.(2023)Hao, Gu, Ma, Hong, Wang, Wang, and Hu]{hao2023reasoning}
Shibo Hao, Yi~Gu, Haodi Ma, Joshua~Jiahua Hong, Zhen Wang, Daisy~Zhe Wang, and Zhiting Hu.
\newblock Reasoning with language model is planning with world model, 2023.

\bibitem[Hendrycks et~al.(2021)Hendrycks, Burns, Kadavath, Arora, Basart, Tang, Song, and Steinhardt]{hendrycksmath2021}
Dan Hendrycks, Collin Burns, Saurav Kadavath, Akul Arora, Steven Basart, Eric Tang, Dawn Song, and Jacob Steinhardt.
\newblock Measuring mathematical problem solving with the math dataset.
\newblock \emph{NeurIPS}, 2021.

\bibitem[Hosseini et~al.(2014)Hosseini, Hajishirzi, Etzioni, and Kushman]{hosseini-etal-2014-learning}
Mohammad~Javad Hosseini, Hannaneh Hajishirzi, Oren Etzioni, and Nate Kushman.
\newblock Learning to solve arithmetic word problems with verb categorization.
\newblock In \emph{Proceedings of the 2014 Conference on Empirical Methods in Natural Language Processing ({EMNLP})}, pp.\  523--533, Doha, Qatar, October 2014. Association for Computational Linguistics.
\newblock \doi{10.3115/v1/D14-1058}.
\newblock URL \url{https://aclanthology.org/D14-1058}.

\bibitem[Kojima et~al.(2023)Kojima, Gu, Reid, Matsuo, and Iwasawa]{kojima2023large}
Takeshi Kojima, Shixiang~Shane Gu, Machel Reid, Yutaka Matsuo, and Yusuke Iwasawa.
\newblock Large language models are zero-shot reasoners, 2023.

\bibitem[Koncel-Kedziorski et~al.(2015)Koncel-Kedziorski, Hajishirzi, Sabharwal, Etzioni, and Ang]{koncel-kedziorski-etal-2015-parsing}
Rik Koncel-Kedziorski, Hannaneh Hajishirzi, Ashish Sabharwal, Oren Etzioni, and Siena~Dumas Ang.
\newblock Parsing algebraic word problems into equations.
\newblock \emph{Transactions of the Association for Computational Linguistics}, 3:\penalty0 585--597, 2015.
\newblock \doi{10.1162/tacl_a_00160}.
\newblock URL \url{https://aclanthology.org/Q15-1042}.

\bibitem[Lai et~al.(2022)Lai, Li, Wang, Zhang, Zhong, Zettlemoyer, tau Yih, Fried, Wang, and Yu]{Lai2022DS1000}
Yuhang Lai, Chengxi Li, Yiming Wang, Tianyi Zhang, Ruiqi Zhong, Luke Zettlemoyer, Scott~Wen tau Yih, Daniel Fried, Sida Wang, and Tao Yu.
\newblock Ds-1000: A natural and reliable benchmark for data science code generation.
\newblock \emph{ArXiv}, abs/2211.11501, 2022.

\bibitem[Li et~al.(2023)Li, Allal, Zi, Muennighoff, Kocetkov, Mou, Marone, Akiki, Li, Chim, Liu, Zheltonozhskii, Zhuo, Wang, Dehaene, Davaadorj, Lamy-Poirier, Monteiro, Shliazhko, Gontier, Meade, Zebaze, Yee, Umapathi, Zhu, Lipkin, Oblokulov, Wang, Murthy, Stillerman, Patel, Abulkhanov, Zocca, Dey, Zhang, Fahmy, Bhattacharyya, Yu, Singh, Luccioni, Villegas, Kunakov, Zhdanov, Romero, Lee, Timor, Ding, Schlesinger, Schoelkopf, Ebert, Dao, Mishra, Gu, Robinson, Anderson, Dolan-Gavitt, Contractor, Reddy, Fried, Bahdanau, Jernite, Ferrandis, Hughes, Wolf, Guha, von Werra, and de~Vries]{li2023starcoder}
Raymond Li, Loubna~Ben Allal, Yangtian Zi, Niklas Muennighoff, Denis Kocetkov, Chenghao Mou, Marc Marone, Christopher Akiki, Jia Li, Jenny Chim, Qian Liu, Evgenii Zheltonozhskii, Terry~Yue Zhuo, Thomas Wang, Olivier Dehaene, Mishig Davaadorj, Joel Lamy-Poirier, João Monteiro, Oleh Shliazhko, Nicolas Gontier, Nicholas Meade, Armel Zebaze, Ming-Ho Yee, Logesh~Kumar Umapathi, Jian Zhu, Benjamin Lipkin, Muhtasham Oblokulov, Zhiruo Wang, Rudra Murthy, Jason Stillerman, Siva~Sankalp Patel, Dmitry Abulkhanov, Marco Zocca, Manan Dey, Zhihan Zhang, Nour Fahmy, Urvashi Bhattacharyya, Wenhao Yu, Swayam Singh, Sasha Luccioni, Paulo Villegas, Maxim Kunakov, Fedor Zhdanov, Manuel Romero, Tony Lee, Nadav Timor, Jennifer Ding, Claire Schlesinger, Hailey Schoelkopf, Jan Ebert, Tri Dao, Mayank Mishra, Alex Gu, Jennifer Robinson, Carolyn~Jane Anderson, Brendan Dolan-Gavitt, Danish Contractor, Siva Reddy, Daniel Fried, Dzmitry Bahdanau, Yacine Jernite, Carlos~Muñoz Ferrandis, Sean Hughes, Thomas Wolf, Arjun Guha, Leandro von
  Werra, and Harm de~Vries.
\newblock Starcoder: may the source be with you!, 2023.

\bibitem[Lightman et~al.(2023)Lightman, Kosaraju, Burda, Edwards, Baker, Lee, Leike, Schulman, Sutskever, and Cobbe]{lightman2023lets}
Hunter Lightman, Vineet Kosaraju, Yura Burda, Harri Edwards, Bowen Baker, Teddy Lee, Jan Leike, John Schulman, Ilya Sutskever, and Karl Cobbe.
\newblock Let's verify step by step, 2023.

\bibitem[Lu et~al.(2023)Lu, Qiu, Yu, Welleck, and Chang]{lu2023survey}
Pan Lu, Liang Qiu, Wenhao Yu, Sean Welleck, and Kai-Wei Chang.
\newblock A survey of deep learning for mathematical reasoning, 2023.

\bibitem[Luo et~al.(2023)Luo, Sun, Xu, Zhao, Lou, Tao, Geng, Lin, Chen, and Zhang]{luo2023wizardmath}
Haipeng Luo, Qingfeng Sun, Can Xu, Pu~Zhao, Jianguang Lou, Chongyang Tao, Xiubo Geng, Qingwei Lin, Shifeng Chen, and Dongmei Zhang.
\newblock Wizardmath: Empowering mathematical reasoning for large language models via reinforced evol-instruct, 2023.

\bibitem[Madaan et~al.(2023)Madaan, Tandon, Gupta, Hallinan, Gao, Wiegreffe, Alon, Dziri, Prabhumoye, Yang, Welleck, Majumder, Gupta, Yazdanbakhsh, and Clark]{madaan2023selfrefine}
Aman Madaan, Niket Tandon, Prakhar Gupta, Skyler Hallinan, Luyu Gao, Sarah Wiegreffe, Uri Alon, Nouha Dziri, Shrimai Prabhumoye, Yiming Yang, Sean Welleck, Bodhisattwa~Prasad Majumder, Shashank Gupta, Amir Yazdanbakhsh, and Peter Clark.
\newblock Self-refine: Iterative refinement with self-feedback, 2023.

\bibitem[OpenAI(2023)]{openai2023gpt4}
OpenAI.
\newblock Gpt-4 technical report, 2023.

\bibitem[Ouyang et~al.(2022)Ouyang, Wu, Jiang, Almeida, Wainwright, Mishkin, Zhang, Agarwal, Slama, Ray, Schulman, Hilton, Kelton, Miller, Simens, Askell, Welinder, Christiano, Leike, and Lowe]{NEURIPS2022_b1efde53}
Long Ouyang, Jeffrey Wu, Xu~Jiang, Diogo Almeida, Carroll Wainwright, Pamela Mishkin, Chong Zhang, Sandhini Agarwal, Katarina Slama, Alex Ray, John Schulman, Jacob Hilton, Fraser Kelton, Luke Miller, Maddie Simens, Amanda Askell, Peter Welinder, Paul~F Christiano, Jan Leike, and Ryan Lowe.
\newblock Training language models to follow instructions with human feedback.
\newblock In S.~Koyejo, S.~Mohamed, A.~Agarwal, D.~Belgrave, K.~Cho, and A.~Oh (eds.), \emph{Advances in Neural Information Processing Systems}, volume~35, pp.\  27730--27744. Curran Associates, Inc., 2022.
\newblock URL \url{https://proceedings.neurips.cc/paper_files/paper/2022/file/b1efde53be364a73914f58805a001731-Paper-Conference.pdf}.

\bibitem[Pan et~al.(2023)Pan, Saxon, Xu, Nathani, Wang, and Wang]{pan2023automatically}
Liangming Pan, Michael Saxon, Wenda Xu, Deepak Nathani, Xinyi Wang, and William~Yang Wang.
\newblock Automatically correcting large language models: Surveying the landscape of diverse self-correction strategies, 2023.

\bibitem[Penedo et~al.(2023)Penedo, Malartic, Hesslow, Cojocaru, Cappelli, Alobeidli, Pannier, Almazrouei, and Launay]{refinedweb}
Guilherme Penedo, Quentin Malartic, Daniel Hesslow, Ruxandra Cojocaru, Alessandro Cappelli, Hamza Alobeidli, Baptiste Pannier, Ebtesam Almazrouei, and Julien Launay.
\newblock The {R}efined{W}eb dataset for {F}alcon {LLM}: outperforming curated corpora with web data, and web data only.
\newblock \emph{arXiv preprint arXiv:2306.01116}, 2023.
\newblock URL \url{https://arxiv.org/abs/2306.01116}.

\bibitem[Radford et~al.(2019)Radford, Wu, Child, Luan, Amodei, and Sutskever]{radford2019language}
Alec Radford, Jeff Wu, Rewon Child, David Luan, Dario Amodei, and Ilya Sutskever.
\newblock Language models are unsupervised multitask learners.
\newblock 2019.

\bibitem[Roy \& Roth(2015)Roy and Roth]{roy-roth-2015-solving}
Subhro Roy and Dan Roth.
\newblock Solving general arithmetic word problems.
\newblock In \emph{Proceedings of the 2015 Conference on Empirical Methods in Natural Language Processing}, pp.\  1743--1752, Lisbon, Portugal, September 2015. Association for Computational Linguistics.
\newblock \doi{10.18653/v1/D15-1202}.
\newblock URL \url{https://aclanthology.org/D15-1202}.

\bibitem[Rozière et~al.(2023)Rozière, Gehring, Gloeckle, Sootla, Gat, Tan, Adi, Liu, Remez, Rapin, Kozhevnikov, Evtimov, Bitton, Bhatt, Ferrer, Grattafiori, Xiong, Défossez, Copet, Azhar, Touvron, Martin, Usunier, Scialom, and Synnaeve]{rozière2023code}
Baptiste Rozière, Jonas Gehring, Fabian Gloeckle, Sten Sootla, Itai Gat, Xiaoqing~Ellen Tan, Yossi Adi, Jingyu Liu, Tal Remez, Jérémy Rapin, Artyom Kozhevnikov, Ivan Evtimov, Joanna Bitton, Manish Bhatt, Cristian~Canton Ferrer, Aaron Grattafiori, Wenhan Xiong, Alexandre Défossez, Jade Copet, Faisal Azhar, Hugo Touvron, Louis Martin, Nicolas Usunier, Thomas Scialom, and Gabriel Synnaeve.
\newblock Code llama: Open foundation models for code, 2023.

\bibitem[Schulman et~al.(2017)Schulman, Wolski, Dhariwal, Radford, and Klimov]{schulman2017proximal}
John Schulman, Filip Wolski, Prafulla Dhariwal, Alec Radford, and Oleg Klimov.
\newblock Proximal policy optimization algorithms, 2017.

\bibitem[Shinn et~al.(2023)Shinn, Cassano, Labash, Gopinath, Narasimhan, and Yao]{shinn2023reflexion}
Noah Shinn, Federico Cassano, Beck Labash, Ashwin Gopinath, Karthik Narasimhan, and Shunyu Yao.
\newblock Reflexion: Language agents with verbal reinforcement learning, 2023.

\bibitem[Taori et~al.(2023)Taori, Gulrajani, Zhang, Dubois, Li, Guestrin, Liang, and Hashimoto]{alpaca}
Rohan Taori, Ishaan Gulrajani, Tianyi Zhang, Yann Dubois, Xuechen Li, Carlos Guestrin, Percy Liang, and Tatsunori~B. Hashimoto.
\newblock Stanford alpaca: An instruction-following llama model.
\newblock \url{https://github.com/tatsu-lab/stanford_alpaca}, 2023.

\bibitem[Touvron et~al.(2023{\natexlab{a}})Touvron, Lavril, Izacard, Martinet, Lachaux, Lacroix, Rozière, Goyal, Hambro, Azhar, Rodriguez, Joulin, Grave, and Lample]{touvron2023llama}
Hugo Touvron, Thibaut Lavril, Gautier Izacard, Xavier Martinet, Marie-Anne Lachaux, Timothée Lacroix, Baptiste Rozière, Naman Goyal, Eric Hambro, Faisal Azhar, Aurelien Rodriguez, Armand Joulin, Edouard Grave, and Guillaume Lample.
\newblock Llama: Open and efficient foundation language models, 2023{\natexlab{a}}.

\bibitem[Touvron et~al.(2023{\natexlab{b}})Touvron, Martin, Stone, Albert, Almahairi, Babaei, Bashlykov, Batra, Bhargava, Bhosale, Bikel, Blecher, Ferrer, Chen, Cucurull, Esiobu, Fernandes, Fu, Fu, Fuller, Gao, Goswami, Goyal, Hartshorn, Hosseini, Hou, Inan, Kardas, Kerkez, Khabsa, Kloumann, Korenev, Koura, Lachaux, Lavril, Lee, Liskovich, Lu, Mao, Martinet, Mihaylov, Mishra, Molybog, Nie, Poulton, Reizenstein, Rungta, Saladi, Schelten, Silva, Smith, Subramanian, Tan, Tang, Taylor, Williams, Kuan, Xu, Yan, Zarov, Zhang, Fan, Kambadur, Narang, Rodriguez, Stojnic, Edunov, and Scialom]{touvron2023llama2}
Hugo Touvron, Louis Martin, Kevin Stone, Peter Albert, Amjad Almahairi, Yasmine Babaei, Nikolay Bashlykov, Soumya Batra, Prajjwal Bhargava, Shruti Bhosale, Dan Bikel, Lukas Blecher, Cristian~Canton Ferrer, Moya Chen, Guillem Cucurull, David Esiobu, Jude Fernandes, Jeremy Fu, Wenyin Fu, Brian Fuller, Cynthia Gao, Vedanuj Goswami, Naman Goyal, Anthony Hartshorn, Saghar Hosseini, Rui Hou, Hakan Inan, Marcin Kardas, Viktor Kerkez, Madian Khabsa, Isabel Kloumann, Artem Korenev, Punit~Singh Koura, Marie-Anne Lachaux, Thibaut Lavril, Jenya Lee, Diana Liskovich, Yinghai Lu, Yuning Mao, Xavier Martinet, Todor Mihaylov, Pushkar Mishra, Igor Molybog, Yixin Nie, Andrew Poulton, Jeremy Reizenstein, Rashi Rungta, Kalyan Saladi, Alan Schelten, Ruan Silva, Eric~Michael Smith, Ranjan Subramanian, Xiaoqing~Ellen Tan, Binh Tang, Ross Taylor, Adina Williams, Jian~Xiang Kuan, Puxin Xu, Zheng Yan, Iliyan Zarov, Yuchen Zhang, Angela Fan, Melanie Kambadur, Sharan Narang, Aurelien Rodriguez, Robert Stojnic, Sergey Edunov, and Thomas
  Scialom.
\newblock Llama 2: Open foundation and fine-tuned chat models, 2023{\natexlab{b}}.

\bibitem[Uesato et~al.(2022)Uesato, Kushman, Kumar, Song, Siegel, Wang, Creswell, Irving, and Higgins]{uesato2022solving}
Jonathan Uesato, Nate Kushman, Ramana Kumar, Francis Song, Noah Siegel, Lisa Wang, Antonia Creswell, Geoffrey Irving, and Irina Higgins.
\newblock Solving math word problems with process- and outcome-based feedback, 2022.

\bibitem[Wang et~al.(2023{\natexlab{a}})Wang, Peng, Jabbarvand, and Ji]{wang2023leti}
Xingyao Wang, Hao Peng, Reyhaneh Jabbarvand, and Heng Ji.
\newblock Leti: Learning to generate from textual interactions, 2023{\natexlab{a}}.

\bibitem[Wang et~al.(2023{\natexlab{b}})Wang, Wei, Schuurmans, Le, Chi, Narang, Chowdhery, and Zhou]{wang2023selfconsistency}
Xuezhi Wang, Jason Wei, Dale Schuurmans, Quoc Le, Ed~Chi, Sharan Narang, Aakanksha Chowdhery, and Denny Zhou.
\newblock Self-consistency improves chain of thought reasoning in language models, 2023{\natexlab{b}}.

\bibitem[Wang et~al.(2023{\natexlab{c}})Wang, Kordi, Mishra, Liu, Smith, Khashabi, and Hajishirzi]{wang-etal-2023-self-instruct}
Yizhong Wang, Yeganeh Kordi, Swaroop Mishra, Alisa Liu, Noah~A. Smith, Daniel Khashabi, and Hannaneh Hajishirzi.
\newblock Self-instruct: Aligning language models with self-generated instructions.
\newblock In \emph{Proceedings of the 61st Annual Meeting of the Association for Computational Linguistics (Volume 1: Long Papers)}, pp.\  13484--13508, Toronto, Canada, July 2023{\natexlab{c}}. Association for Computational Linguistics.
\newblock \doi{10.18653/v1/2023.acl-long.754}.
\newblock URL \url{https://aclanthology.org/2023.acl-long.754}.

\bibitem[Wei et~al.(2023)Wei, Wang, Schuurmans, Bosma, Ichter, Xia, Chi, Le, and Zhou]{wei2023chainofthought}
Jason Wei, Xuezhi Wang, Dale Schuurmans, Maarten Bosma, Brian Ichter, Fei Xia, Ed~Chi, Quoc Le, and Denny Zhou.
\newblock Chain-of-thought prompting elicits reasoning in large language models, 2023.

\bibitem[Weng et~al.(2023)Weng, Zhu, Xia, Li, He, Liu, and Zhao]{weng2023large}
Yixuan Weng, Minjun Zhu, Fei Xia, Bin Li, Shizhu He, Kang Liu, and Jun Zhao.
\newblock Large language models are better reasoners with self-verification, 2023.

\bibitem[Wu et~al.(2023)Wu, Hu, Shi, Dziri, Suhr, Ammanabrolu, Smith, Ostendorf, and Hajishirzi]{wu2023finegrained}
Zeqiu Wu, Yushi Hu, Weijia Shi, Nouha Dziri, Alane Suhr, Prithviraj Ammanabrolu, Noah~A. Smith, Mari Ostendorf, and Hannaneh Hajishirzi.
\newblock Fine-grained human feedback gives better rewards for language model training, 2023.

\bibitem[Xie et~al.(2023)Xie, Kawaguchi, Zhao, Zhao, Kan, He, and Xie]{xie2023decomposition}
Yuxi Xie, Kenji Kawaguchi, Yiran Zhao, Xu~Zhao, Min-Yen Kan, Junxian He, and Qizhe Xie.
\newblock Decomposition enhances reasoning via self-evaluation guided decoding, 2023.

\bibitem[Yang et~al.(2023)Yang, Swope, Gu, Chalamala, Song, Yu, Godil, Prenger, and Anandkumar]{yang2023leandojo}
Kaiyu Yang, Aidan Swope, Alex Gu, Rahul Chalamala, Peiyang Song, Shixing Yu, Saad Godil, Ryan Prenger, and Anima Anandkumar.
\newblock {LeanDojo}: Theorem proving with retrieval-augmented language models.
\newblock \emph{arXiv preprint arXiv:2306.15626}, 2023.

\bibitem[Yao et~al.(2023{\natexlab{a}})Yao, Yu, Zhao, Shafran, Griffiths, Cao, and Narasimhan]{yao2023tree}
Shunyu Yao, Dian Yu, Jeffrey Zhao, Izhak Shafran, Thomas~L. Griffiths, Yuan Cao, and Karthik Narasimhan.
\newblock Tree of thoughts: Deliberate problem solving with large language models, 2023{\natexlab{a}}.

\bibitem[Yao et~al.(2023{\natexlab{b}})Yao, Zhao, Yu, Du, Shafran, Narasimhan, and Cao]{yao2023react}
Shunyu Yao, Jeffrey Zhao, Dian Yu, Nan Du, Izhak Shafran, Karthik Narasimhan, and Yuan Cao.
\newblock React: Synergizing reasoning and acting in language models, 2023{\natexlab{b}}.

\bibitem[Yuan et~al.(2023)Yuan, Yuan, Tan, Wang, Huang, and Huang]{yuan2023rrhf}
Zheng Yuan, Hongyi Yuan, Chuanqi Tan, Wei Wang, Songfang Huang, and Fei Huang.
\newblock Rrhf: Rank responses to align language models with human feedback without tears, 2023.

\bibitem[Zhang et~al.(2022{\natexlab{a}})Zhang, Roller, Goyal, Artetxe, Chen, Chen, Dewan, Diab, Li, Lin, Mihaylov, Ott, Shleifer, Shuster, Simig, Koura, Sridhar, Wang, and Zettlemoyer]{zhang2022opt}
Susan Zhang, Stephen Roller, Naman Goyal, Mikel Artetxe, Moya Chen, Shuohui Chen, Christopher Dewan, Mona Diab, Xian Li, Xi~Victoria Lin, Todor Mihaylov, Myle Ott, Sam Shleifer, Kurt Shuster, Daniel Simig, Punit~Singh Koura, Anjali Sridhar, Tianlu Wang, and Luke Zettlemoyer.
\newblock Opt: Open pre-trained transformer language models, 2022{\natexlab{a}}.

\bibitem[Zhang et~al.(2022{\natexlab{b}})Zhang, Zhang, Li, and Smola]{zhang2022automatic}
Zhuosheng Zhang, Aston Zhang, Mu~Li, and Alex Smola.
\newblock Automatic chain of thought prompting in large language models, 2022{\natexlab{b}}.

\bibitem[Zhao et~al.(2023)Zhao, Xie, Kawaguchi, He, and Xie]{zhao2023automatic}
Xu~Zhao, Yuxi Xie, Kenji Kawaguchi, Junxian He, and Qizhe Xie.
\newblock Automatic model selection with large language models for reasoning, 2023.

\end{thebibliography}
